\documentclass[sigconf,nonacm]{acmart}
\usepackage{tikz}
\usepackage{placeins}
\usepackage{array}
\usepackage{float}
\usepackage{algorithm}
\usepackage{algpseudocode}
\usetikzlibrary{arrows.meta,positioning,fit,calc}

\setcopyright{none}
\newcommand{\NeMoGym}{\href{https://github.com/NVIDIA-NeMo/Gym}{\textcolor{blue}{NeMo Gym}}}

\title[Trace2Skill]{Trace2Skill: Verifier-Guided Skill Evolution for Long-Context EDA Agents}

\author{Zijian Du}
\affiliation{%
  \institution{NVIDIA}
  \city{Santa Clara}
  \state{CA}
  \country{USA}}
\email{ldu@nvidia.com}

\author{Nathaniel Pinckney}
\affiliation{%
  \institution{NVIDIA}
  \city{Santa Clara}
  \state{CA}
  \country{USA}}
\email{npinckney@nvidia.com}

\begin{document}
\raggedbottom

\begin{abstract}
Complex Verilog Design Problems (CVDP) challenge hardware LLM agents because solving them requires localizing verifier-relevant RTL, testbenches, include paths, and build dependencies inside large repository snapshots, making precise edits, and recovering from sparse hidden-verifier failures. We present Trace2Skill, a test-time scaling framework that improves a hardware agent without RTL-specialized model fine-tuning. Rather than training a new model or only sampling more candidate solutions, Trace2Skill treats the agent's natural-language skill as an evolvable policy. It mines repeated rollout traces for both success modes and failure modes, converts them into dense diagnostics and oracle lessons, and uses an oracle--mutator--selector loop to produce task-specific skills that guide later search, editing, validation, and recovery. Because final pass/fail labels are often too coarse for hard failures, Trace2Skill also supports bounded runtime dense verifier feedback that returns sanitized functional observations while keeping hidden harnesses and reference solutions inaccessible to the agent. This feedback helps better guide skill evolution and agent execution by tightening a co-optimization loop among skill text, verifier evidence, and downstream behavior. Across hard CVDP tasks that defeat the seed CVDP agent, including tasks that also defeat frontier coding agents, Trace2Skill with dense verifier feedback substantially improves task pass rates and produces breakthrough passes on previously unsolved tasks, without requiring high-quality fine-tuning data, specialized RTL model training, or model weight updates. The same framework provides a general test-time scaling strategy that can extend beyond digital design to other verifiable EDA tasks.

\end{abstract}


\maketitle

\begin{figure*}[t]
\centering
\resizebox{0.61\textwidth}{!}{%
\begin{tikzpicture}[
  font=\fontsize{12}{13}\selectfont,
  stage/.style={rectangle, rounded corners=2pt, draw=black!70, line width=1.05pt,
    align=center, text width=4.12cm, minimum height=1.22cm, inner sep=4pt},
  smallstage/.style={rectangle, rounded corners=2pt, draw=black!65, line width=1.0pt,
    align=center, text width=3.95cm, minimum height=0.88cm, inner sep=3pt},
  densebox/.style={smallstage, dashed, draw=black!70, line width=1.05pt},
  wide/.style={rectangle, rounded corners=2pt, draw=black!60, line width=1.0pt,
    align=center, text width=10.2cm, minimum height=0.82cm, inner sep=3pt},
  arrow/.style={-{Latex[length=1.9mm]}, thick, draw=black!75},
  looparrow/.style={-{Latex[length=2.2mm]}, line width=1.45pt, draw=black!85},
  densearrow/.style={Latex-Latex, thick, dashed, draw=black!70}
]
  \definecolor{taskc}{RGB}{214,232,255}
  \definecolor{rollc}{RGB}{217,244,226}
  \definecolor{evalc}{RGB}{255,236,178}
  \definecolor{skillc}{RGB}{232,222,248}
  \definecolor{densec}{RGB}{255,218,205}
  \definecolor{telec}{RGB}{226,226,226}

  \node[stage, fill=taskc] (task) at (0,0) {1. Task\\repo + prompt};
  \node[stage, fill=skillc] (skill) at (4.65,0) {2. Candidate skills\\seed, survivor, children};
  \node[stage, fill=rollc] (rollout) at (9.3,0) {3. Rollouts\\tool agent, repeats\\{\fontsize{7.5}{8.5}\selectfont\bfseries Claude Opus 4.6}};

  \node[stage, fill=evalc] (aggregate) at (0,-1.65) {6. Select survivor\\+ aggregate evidence};
  \node[stage, fill=evalc] (fitness) at (4.65,-1.65) {5. Metrics\\PassRate + SelectQ};
  \node[stage, fill=evalc] (verify) at (9.3,-1.65) {4. Final verify\\hidden tests};

  \node[stage, fill=skillc] (oracle) at (0,-3.3) {7. Oracle\\trace lessons\\{\fontsize{7.5}{8.5}\selectfont\bfseries GPT-5}};
  \node[stage, fill=skillc] (mutator) at (4.65,-3.3) {8. Mutate + repair\\next children\\{\fontsize{7.5}{8.5}\selectfont\bfseries Claude Sonnet 4.5}};
  \node[stage, fill=densec] (select) at (9.3,-3.3) {9. Next population\\carry survivor + children};

  \node[densebox, fill=densec] (dense) at (9.3,1.85) {Runtime\\feedback};
  \node[wide, fill=telec] (logs) at (4.65,-4.82) {Traceability artifacts: rollouts, metrics, feedback, skills, survivor decisions};

  \node[rectangle, rounded corners=1pt, draw=black!35, fill=white,
    minimum width=10.65cm, minimum height=1.12cm] at (4.65,-6.12) {};
  \node[rectangle, draw=black!45, fill=taskc, minimum width=0.38cm, minimum height=0.22cm] at (0.15,-5.86) {};
  \node[anchor=west, font=\fontsize{10.2}{11.0}\selectfont] at (0.42,-5.86) {task};
  \node[rectangle, draw=black!45, fill=skillc, minimum width=0.38cm, minimum height=0.22cm] at (3.95,-5.86) {};
  \node[anchor=west, font=\fontsize{10.2}{11.0}\selectfont] at (4.22,-5.86) {skill/learning};
  \node[rectangle, draw=black!45, fill=rollc, minimum width=0.38cm, minimum height=0.22cm] at (7.55,-5.86) {};
  \node[anchor=west, font=\fontsize{10.2}{11.0}\selectfont] at (7.82,-5.86) {rollout};
  \node[rectangle, draw=black!45, fill=evalc, minimum width=0.38cm, minimum height=0.22cm] at (0.15,-6.32) {};
  \node[anchor=west, font=\fontsize{10.2}{11.0}\selectfont] at (0.42,-6.32) {evaluation};
  \node[rectangle, draw=black!45, fill=densec, minimum width=0.38cm, minimum height=0.22cm] at (3.95,-6.32) {};
  \node[anchor=west, font=\fontsize{10.2}{11.0}\selectfont] at (4.22,-6.32) {feedback/selection};
  \node[rectangle, draw=black!45, fill=telec, minimum width=0.38cm, minimum height=0.22cm] at (7.55,-6.32) {};
  \node[anchor=west, font=\fontsize{10.2}{11.0}\selectfont] at (7.82,-6.32) {artifacts};

  \draw[arrow] (task) -- (skill);
  \draw[arrow] (skill) -- (rollout);
  \draw[arrow] (rollout) -- (verify.north);
  \draw[arrow] (verify) -- (fitness);
  \draw[arrow] (fitness) -- (aggregate);
  \draw[arrow] (aggregate) -- (oracle.north);
  \draw[arrow] (oracle) -- (mutator);
  \draw[arrow] (mutator) -- (select);
  \coordinate (loopright) at ($(select.east)+(0.35,0)$);
  \coordinate (looptop) at ($(skill.north)+(0,0.28)$);
  \draw[looparrow] (select.east) -- (loopright) |- (looptop) -- (skill.north);
  \draw[densearrow] (rollout.north) -- (dense.south);
\end{tikzpicture}
}
\caption{Trace2Skill end-to-end flow. Colors group component roles as shown in the legend; solid arrows mark the main execution/evaluation path, the thick return arrow marks skill evolution into the next population, and the dashed bidirectional edge marks optional runtime feedback. Model labels identify LLM-driven stages.}
\Description{A flow diagram showing task packaging, candidate-skill injection, Claude Opus
rollouts, optional runtime verifier feedback, hidden verification, dense metrics, survivor
selection with per-task aggregation, GPT-5 oracle feedback, Claude Sonnet mutation and repair,
next-population construction, and trace logging.}
\label{fig:method-flow}
\end{figure*}
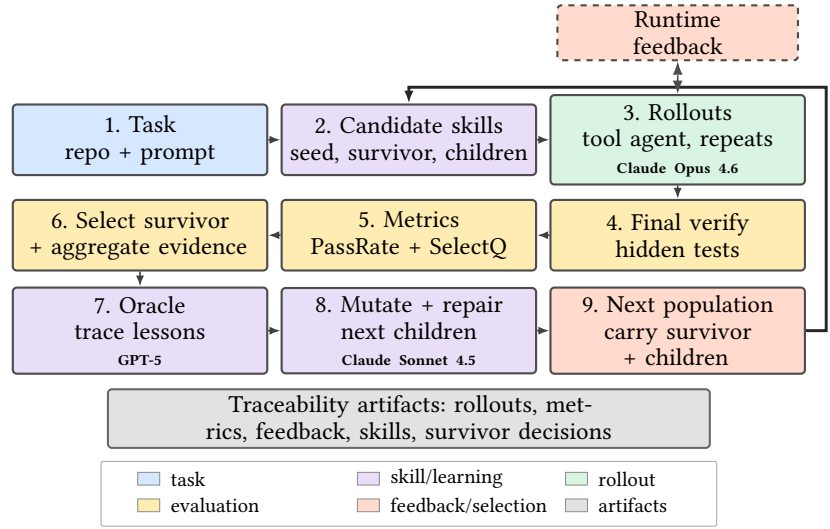

\section{Introduction}

Large language models are moving from passive text generators to tool-using agents that reason, inspect state, act, and revise their work. Recent systems interleave reasoning with actions \cite{yao2023react}, learn from execution feedback without weight updates \cite{shinn2023reflexion}, accumulate reusable skills over long horizons \cite{wang2023voyager}, and expose agent-computer interfaces for software engineering \cite{yang2024sweagent}. Multi-agent and benchmark work further shows that tool use, role structure, and realistic interactive environments are becoming central to agent research \cite{wu2023autogen,hong2023metagpt,liu2023agentbench,zhou2023webarena}.

Hardware design challenges this progress. Prior work has advanced RTL/code-generation benchmarks \cite{liu2023verilogeval,lu2024rtllm,batten2024pyhdl,pinckney2025cvdp}, chip-domain alignment and reasoning adaptation \cite{deng2025chipalign,deng2025scalertl}, and compiler-, simulator-, retrieval-, or waveform-guided Verilog agents \cite{tsai2024rtlfixer,ho2025verilogcoder}. Most closely related, ACE-RTL combines an RTL-specialized generator trained on large-scale RTL data with frontier-model reflection, per-problem context evolution, and parallel test-time scaling to obtain strong agentic pass rates on full CVDP categories; it reports that parallel scaling reduces average iterations to first success from 11.30 to 4.08 across four CVDP categories \cite{deng2026acertl}. These systems show clear promise, but CVDP also exposes a harder recovery regime: long-context, multi-turn tasks that require repository navigation, bug localization, feature integration, RTL editing, tool invocation, and recovery from failed verification. Many such tasks fail frontier agents and provide little learning signal for agents to improve.

Trace2Skill studies test-time scaling for this recovery setting. Instead of fine-tuning model weights, it evolves the natural-language skill injected into a fixed CVDP agent. The skill is a policy artifact that guides search, editing, validation, and recovery. After each rollout batch, Trace2Skill mines execution traces for pass and failure modes, asks an oracle to summarize actionable lessons, and uses an evolutionary mutator and selector to produce the next task-specific skill. Runtime dense verifier feedback is an optional black-box tool that provides additional learning signal during rollout: it returns sanitized functional observations while keeping hidden harnesses and reference solutions inaccessible. This can make failures more actionable, but may incur additional overhead when verification is compute-heavy.

Our goal is therefore complementary to ACE-RTL~\cite{deng2026acertl}. ACE-RTL primarily asks how far a trained RTL-specialized model and evolving per-problem context can lift pass rates across broad CVDP categories. Trace2Skill asks a different question: when the seed CVDP agent fails, can execution traces from an open agent mine useful skills without model fine-tuning, including on tasks that also defeat frontier coding agents? The unit of adaptation is not only the current RTL candidate or problem context, but the reusable skill document that governs future rollouts; the evaluation correspondingly focuses on hard-failure recovery, dense behavioral credit assignment, and the relationship between learned skill content and agent execution quality.

These observations shape our design. Frontier coding agents are strong black-box baselines, but the open CVDP agent exposes traces, verifier outcomes, and skill text, making task-level adaptation interpretable and tunable. Because sparse pass/fail labels are too coarse for hard failures, we design granular metrics covering both skill quality and agent execution quality as proxy optimization targets, and show that they correlate highly with sparse pass/fail signals. Finally, hard CVDP tasks are heterogeneous: shared skills transfer unreliably, motivating task-wise evolution rather than multi-task co-optimization.

This paper makes the following contributions:
\begin{list}{\labelitemi}{%
  \setlength{\leftmargin}{1.08em}%
  \setlength{\labelwidth}{0.55em}%
  \setlength{\labelsep}{0.35em}%
  \setlength{\rightmargin}{0pt}%
  \setlength{\topsep}{0.15em}%
  \setlength{\partopsep}{0pt}%
  \setlength{\parsep}{0pt}%
  \setlength{\itemsep}{0.18em}%
}
  \item \textbf{Trace2Skill, a training-free framework that converts CVDP execution traces into task-specific agent skills without model weight updates.}
  \item \textbf{A decomposed dense quality signal that ranks failed rollouts by separating skill content from agent execution behavior.}
  \item \textbf{An open, interpretable CVDP Trace2Skill agent that solves hard tasks that fail the baseline CVDP agent, including tasks that also fail Claude Code and Codex.}
\end{list}

\section{Method}

Figure~\ref{fig:method-flow} gives the execution loop used by all configurations. Deterministic harness components load tasks, inject skills, enforce runtime-feedback policy, run hidden verification, compute metrics, aggregate per-task evidence, and record artifacts; LLM-driven blocks perform rollout, oracle summarization, and skill mutation.

\subsection{Execution and Feedback}
Each CVDP task provides a repository snapshot and objective. Baseline rollouts inject a CID-specific seed skill; optimization rollouts inject the current task-specific survivor skill. We run these tool-agent rollouts through \NeMoGym{}: the head server launches submitted runs, the agent alternates model calls with resource-server tool execution, and the resource server returns tool observations and final verifier rewards, as shown in Appendix Figure~\ref{fig:app-nemogym-rollout}. A Claude Opus~4.6 tool agent then inspects files, edits RTL, compiles, simulates, and stops after a 30-turn budget or completion. The hidden verifier gives the official final pass/fail label. In dense-feedback configurations, the agent may also call \texttt{verify\_feedback}: a budgeted black-box verifier run on a private workspace copy. It returns sanitized functional feedback while hiding reference solutions, hidden harness files, raw cocotb source/logs, hidden paths, and Docker configuration.

\subsection{Metrics}
\label{sec:learning-signal}
\label{sec:diagnostic-metrics}

Agents frequently fail the hard long-context CVDP tasks, leading many rollouts to score zero. We
therefore use final verifier pass rate as the final goal, but use carefully
designed dense signals to drive a more granular learning process. Component dense
scores are clipped to \([0,1]\), and the same weights are used across tasks.

\begin{table*}[t]
\centering
\caption{Four-configuration ablation on the 8 hard tasks with zero CVDP baseline agent passes. All conditions use the same task set, rollout model, 30-turn cap, and final hidden verifier.}
\label{tab:four-config-doe}
\footnotesize
\renewcommand{\arraystretch}{1.12}
\setlength{\tabcolsep}{2.5pt}
\begin{tabular}{@{}>{\raggedright\arraybackslash}p{0.045\textwidth}>{\raggedright\arraybackslash}p{0.12\textwidth}>{\raggedright\arraybackslash}p{0.18\textwidth}>{\raggedright\arraybackslash}p{0.205\textwidth}>{\raggedright\arraybackslash}p{0.19\textwidth}>{\raggedright\arraybackslash}p{0.17\textwidth}@{}}
\toprule
\textbf{Config} & \textbf{Agent} & \textbf{EA schedule} & \textbf{Runtime feedback} & \textbf{Budget} & \textbf{Purpose} \\
\midrule
C1 & CVDP baseline & None & Final sparse verifier only & Eval: $8\times4=32$ rollouts & Seed-skill baseline. \\
C2 & CVDP baseline & None & Dense feedback ($\leq3$ calls/rollout) plus final verifier & Same as C1 & Seed skill with dense feedback. \\
C3 & CVDP Trace2Skill & $5$ EA generations; $4$ skills/gen.; $4$ rollout repeats/skill & Same as C1 & Optimization: $8\times5\times4\times4=640$ rollouts & Skill evolution under sparse reward. \\
C4 & CVDP Trace2Skill & Same as C3 & Same as C2 & Same as C3 & Skill evolution under dense reward. \\
\midrule
\multicolumn{6}{@{}l@{}}{\textit{Unified outcome metrics:} pass rate over attempted rollouts, tasks solved/8, and AgentProgressQ.} \\
\bottomrule
\end{tabular}
\end{table*}

\subsubsection{PassRate: final correctness}
\begin{equation}
\mathrm{PassRate}=\frac{1}{R}\sum_{r=1}^{R}p_r .
\end{equation}
Here \(p_r\in\{0,1\}\) is the final hidden-verifier result for repeat \(r\): \(p_r=0\) means the agent submitted a final solution that failed verification, while \(p_r=1\) means its submitted solution passed the hidden verifier.
Runtime dense feedback, when enabled, never counts as final correctness; it only
helps the agent choose later edits.

\smallskip
\refstepcounter{subsubsection}
\noindent\textbf{\thesubsubsection}\quad\textit{SkillQ: skill content quality.}\par\noindent
SkillQ scores the repaired skill text before future rollouts. It measures
whether a child skill preserves parent guidance and integrates oracle lessons in
executable, agent-visible form:
\begin{equation}
\begin{aligned}
Q_{\mathrm{skill}}^{\mathrm{raw}}={}&0.35L+0.30G+0.10R_p\\
&+0.15A_{\mathrm{act}}+0.05V_s+0.03N+0.02D .
\end{aligned}
\end{equation}
\begin{equation}
Q_{\mathrm{skill}}=Q_{\mathrm{skill}}^{\mathrm{raw}}(0.55+0.45V_s)M_{\mathrm{keep}}.
\end{equation}
Here \(Q_{\mathrm{skill}}^{\mathrm{raw}}\) is the ungated content score; \(L\) is critical lesson coverage, \(G\) evidence grounding, \(R_p\) parent-rule
retention, \(A_{\mathrm{act}}\) actionability, \(V_s\) safety validity, \(N\)
non-redundancy, and \(D\) conservative mutation size. \(M_{\mathrm{keep}}\) is a
retention gate: it penalizes a proposal that drops critical oracle directives
learned from previous rollouts. Hidden-harness leakage, unavailable-tool advice,
contradictions, and semantic-floor regressions are repaired; if still present,
the candidate cannot survive. Overall, SkillQ favors children that add grounded
lessons from the current generation while preserving previous-generation
guidance and excluding harmful or rule-violating advice.

\smallskip
\refstepcounter{subsubsection}
\noindent\textbf{\thesubsubsection}\quad\textit{AgentProgressQ: rollout execution quality.}\par\noindent
AgentProgressQ measures what the agent actually accomplished during a rollout,
even when the final verifier fails. It does not score the skill text directly;
instead, it measures the agent behavior guided by the injected skill:
\begin{equation}
F_{\mathrm{base}}=0.40V+0.20X+0.15H+0.15E+0.10\eta .
\end{equation}
\begin{equation}
F_{\mathrm{progress}}=F_{\mathrm{base}}(0.55+0.45P_{\mathrm{path}}).
\end{equation}
The terms are: \(V\), verifier or partial-verifier progress; \(X\), execution
phase reached, from inspection through edit/compile/simulation; \(H\), alignment
with visible harness conventions; \(E\), edit quality; \(\eta\), turn efficiency;
and \(P_{\mathrm{path}}\), a penalty for wrong-target edits or validation. We observed undesirable stochastic behavior across repeated rollouts under the
same skill, so repeated measurements should penalize unstable progress. For
repeated rollouts under the same candidate skill, we use a lower-confidence-bound
summary:
\begin{equation}
F_{\mathrm{LCB}}=\max\left(0,\bar F_{\mathrm{progress}}-1.96\sigma/\sqrt{R}\right).
\end{equation}
\begin{equation}
Q_{\mathrm{agent}}^{\mathrm{prog}}=0.80F_{\mathrm{LCB}}+0.20\bar F_{\mathrm{progress}}.
\end{equation}
Here \(R\) is the number of repeated rollouts for the candidate skill and \(\sigma\) is the standard deviation of their \(F_{\mathrm{progress}}\) scores. The LCB term reduces the chance that a single lucky rollout dominates selection. Statistically, the LCB term estimates uncertainty in the mean progress, so its penalty scales as \(\sigma/\sqrt{R}\). AgentVarianceQ below instead characterizes rollout instability itself: AgentProgressQ penalizes uncertainty inside the optimization signal, while AgentVarianceQ reports how stochastic the rollout behavior was.

\subsubsection{AgentVarianceQ: repeat stability}
\begin{equation}
Q_{\mathrm{agent}}^{\mathrm{var}}=1-\min(1,\sigma/0.30).
\end{equation}
This diagnostic uses only rollout behavior. High values mean repeated attempts
under the same skill behave consistently; low values warn that generation-level
means may reflect stochastic long-horizon variation. When pass rates are tied,
a higher AgentVarianceQ indicates greater behavioral consistency and is therefore
preferred as a diagnostic selection signal.

\smallskip
\refstepcounter{subsubsection}
\noindent\textbf{\thesubsubsection}\quad\textit{SelectQ: evolutionary survivor score.}\par\noindent
SelectQ is the score used to choose the next parent skill:
\begin{equation}
\mathrm{SelectQ}(S)=
\begin{cases}
-1, & B(S)=1,\\
\mathrm{PassRate}(S)+\epsilon U(S), & B(S)=0,
\end{cases}
\end{equation}
\begin{equation}
U=0.60F_{\mathrm{LCB}}+0.20\bar F_{\mathrm{progress}}+0.20Q_{\mathrm{skill}} .
\end{equation}
\begin{equation}
\epsilon=0.49 / \max(R,N_{\mathrm{task}},1) .
\end{equation}
\(B(S)=1\) marks an invalid or semantically regressed skill. The small
\(\epsilon\) makes SelectQ pass-dominant: one additional hidden-verifier pass
always outweighs dense tie-breakers, while AgentProgressQ and SkillQ rank
candidates with the same pass rate. Although SkillQ and the agent-quality terms
are measured separately, SelectQ combines them because skill content and rollout
behavior are coupled: a semantically plausible skill is useful only if it induces
better agent execution. Empirically, selecting only by SkillQ can preserve clean
but behaviorally stagnant guidance, while selecting only by AgentProgressQ can
chase noisy rollout outcomes. SelectQ therefore combines semantic skill quality
with repeat-robust agent progress, requiring valid guidance while favoring skills
that improve execution.

\subsection{Skill Evolution and Traceability}
Trace2Skill aggregates evidence only within the current target task. A GPT-5 oracle converts the task-local rollouts into lessons about useful behaviors and repeated failures. A Claude Sonnet~4.5 mutator proposes child skills from the survivor skill, oracle lessons, cumulative lesson bank, and tool/visibility contract. Deterministic post-processing repairs or salvages useful proposals while removing unsafe or ungrounded guidance. The next generation evaluates the carried parent and repaired children under the same rollout protocol; Appendix Algorithm~\ref{alg:appendix-parent-child-flow} gives the formal update. All stages write artifacts: raw traces, verifier outputs, feedback calls, dense metrics, oracle summaries, mutator proposals, repaired skills, and survivor decisions.

\section{Experimental Setup}

First, we establish a CVDP baseline agent using unoptimized CID seed skills on 24 public, OSS-runnable long-context CVDP tasks and compare it with Codex and Claude Code under the same hidden CVDP verifier. The 24 tasks include six each from cid003 (specification-to-RTL), cid004 (RTL refinement), cid005 (code completion), and cid016 (RTL bug fixing); they avoid commercial EDA tools and use Docker, Icarus Verilog (\texttt{iverilog}/\texttt{vvp}), verifier images compatible with cocotb, and the CVDP resource server. Second, we select the eight seed-CVDP failures shown in bold in Appendix Table~\ref{tab:oss24-task-outcomes} and study whether task-wise Trace2Skill evolution and runtime dense verifier feedback improve the CVDP agent on those hard failures. The same appendix table records which of these tasks were also failed by Claude Code and Codex.

\vspace*{-2.0em}
\subsection{Configurations}
The 24-task baseline compares the unoptimized seed-skill CVDP agent, Claude Code, and native Codex under the same hidden CVDP verifier. The CVDP agent uses the CVDP harness with fixed CID seed skills and without EA or dense feedback; frontier agents use their own harnesses and receive none of our skills. For the 8-task hard subset, Table~\ref{tab:four-config-doe} shows the four configurations we ablate: C1 is the failed CVDP baseline agent, C2 adds dense feedback, C3 adds task-wise Trace2Skill with sparse final reward, and C4 combines task-wise Trace2Skill with dense feedback.

CVDP rollouts use Claude Opus~4.6 at temperature 0.2. Oracle summarization uses GPT-5 at temperature 0.0, and mutation uses Claude Sonnet~4.5 at temperature 0.35 followed by deterministic repair. C1/C2 use four repeats per task, giving 32 baseline attempts. C3/C4 use five generations, four candidate skills per generation, and four repeats per candidate: 80 optimization rollouts per task and 640 per EA configuration. We report C3/C4 from the optimization rollouts themselves, because this study targets task-wise adaptation on the same hard tasks rather than train/test skill generalization.

Dense-feedback runs expose \texttt{verify\_feedback(reason)} at most three times per rollout, only after a visible code edit, successful visible compile, and a code change since the previous call. The returned object contains sanitized pass/fail, partial test count, failure phase, and a short next-focus hint. Regression tests and telemetry check tool exposure, call-budget enforcement, sanitization, and final verifier outcomes.

\vspace*{-1.0em}
\subsection{AgentQ and Pass/Fail Correlation}
\label{sec:agent-progress-pass-correlation}
Before using AgentQ as an EA tie-breaker, we check whether it aligns with final sparse success on seed-skill baseline rollouts. Because this calibration is run on the CVDP agent without EA, it uses agent-behavior progress only: \(F_{\mathrm{progress}}\) for individual rollouts and \(Q_{\mathrm{agent}}^{\mathrm{prog}}\) for repeated measurements. The completed 96 baseline rollouts give a high point-biserial correlation (\(r=0.90\)) and AUC of 0.982, indicating that the proxy tracks sparse pass/fail outcomes closely enough to serve as an optimization target when hard tasks otherwise provide identical zero-pass rewards.

Figure~\ref{fig:dense-proxy} tests whether the proxy points toward true pass/fail behavior; it does not replace hidden verification, which remains the leading term in SelectQ, but it supplies differential guidance that can move the agent toward eventual verifier passes.

{\setlength{\intextsep}{0.15em}%
\begin{figure}[H]
  \centering
  \includegraphics[width=0.72\linewidth]{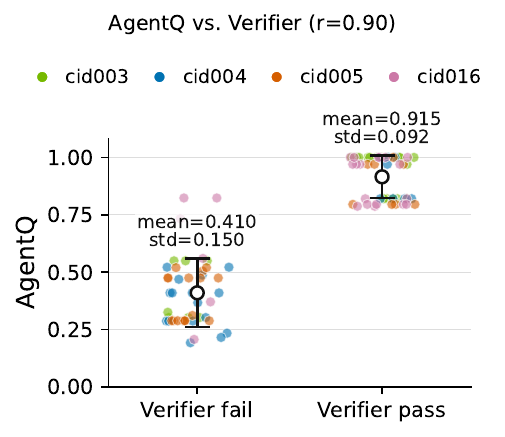}
  \caption{AgentQ proxy on 96 completed OSS seed-skill baseline rollouts, grouped by hidden verifier outcome and colored by CID. Each point is one rollout.}
  \Description{Scatter plot comparing AgentQ for verifier-failed and verifier-passed OSS24 rollouts. Points are color-coded by CVDP CID, with black mean markers and one-standard-deviation error bars.}
  \label{fig:dense-proxy}
\end{figure}}

\FloatBarrier
\section{Results and Analysis}\label{sec:results-analysis}

\subsection{Baseline Agent Comparison}
We first evaluate the matched public OSS CVDP subset: six tasks each from cid003, cid004, cid005, and cid016. This establishes the unoptimized CVDP agent baseline before any Trace2Skill adaptation.

\begin{table}[H]
\centering
\caption{Matched public OSS CVDP baseline results by CID. The subset contains
public long-context agentic CVDP tasks whose harnesses do not require commercial
EDA tools, with six sampled tasks per OSS-runnable CID.}
\label{tab:public-oss-seed-vs-claude}
\small
\setlength{\tabcolsep}{3pt}
\begin{tabular}{@{}llccc@{}}
\toprule
CID & Category & CVDP agent & Claude Code & Codex \\
\midrule
cid003 & Spec2RTL & 4/6 (66.7\%) & 5/6 (83.3\%) & 4/6 (66.7\%) \\
cid004 & RTL refine & 2/6 (33.3\%) & 3/6 (50.0\%) & 2/6 (33.3\%) \\
cid005 & Complete & 1/6 (16.7\%) & 2/6 (33.3\%) & 2/6 (33.3\%) \\
cid016 & Bug fix & 2/6 (33.3\%) & 2/6 (33.3\%) & 4/6 (66.7\%) \\
\midrule
\textbf{Overall} & \textbf{OSS subset} & \textbf{9/24 (37.5\%)} & \textbf{12/24 (50.0\%)} & \textbf{12/24 (50.0\%)} \\
\bottomrule
\end{tabular}
\end{table}

Table~\ref{tab:public-oss-seed-vs-claude} shows that the seed-skill CVDP agent solves 9/24 tasks, while Claude Code and Codex each solve 12/24. The frontier agents are stronger black-box pass-rate baselines, but the CVDP agent exposes the traces, skills, tools, and verifier outcomes needed for test-time scaling. The CID breakdown is also informative: aggregated across the three agents, Spec2RTL has the highest solve rate (13/18), while bug fixing (8/18), RTL refinement (7/18), and code completion (5/18) are lower. This suggests that the agents already have substantial RTL programming ability, but performance drops when success requires long-context repository localization, editing existing RTL, debugging failures, and preserving compatibility with surrounding code. The bold tasks in Appendix Table~\ref{tab:oss24-task-outcomes} are the eight seed-CVDP failures used for the C1--C4 ablations; the same table shows that this set includes both all-agent failures and tasks that frontier agents solved without our trace-adaptation substrate.

\subsection{Final Evaluation on Hard Tasks}
We next evaluate whether test-time scaling can solve the eight hard tasks that the seed CVDP agent failed. C1 is the seed-skill CVDP agent with only final sparse verification. C2 adds runtime dense verifier feedback but no skill evolution. C3 adds task-wise Trace2Skill EA under sparse final reward. C4 combines the same EA budget with runtime dense verifier feedback. This is not a train/test generalization experiment: prior shared-skill tests suggested weak transfer across heterogeneous tasks, so each condition is evaluated on the same hard-task set.

\begin{table}[!htbp]
\centering
\caption{C1--C4 comparison on the same 8 hard tasks.}
\label{tab:c1-c4-final}
\small
\renewcommand{\arraystretch}{1.18}
\setlength{\tabcolsep}{3.6pt}
\begin{tabular}{@{}lccrrrr@{}}
\toprule
\textbf{Cfg.} & \textbf{EA} & \textbf{Dense} & \textbf{Rollouts} & \textbf{Pass Rate} & \textbf{Solved/8} & \textbf{AgentQ} \\
\midrule
C1 & No & No & 32 & 0.0\% & 0 & 0.34 \\
C2 & No & Yes & 32 & 12.5\% & 2 & 0.42 \\
C3 & Yes & No & 640 & 27.0\% & 3 & 0.523 \\
C4 & Yes & Yes & 640 & 33.6\% & 6 & 0.566 \\
\bottomrule
\end{tabular}
\end{table}

Table~\ref{tab:c1-c4-final} gives the final outcome view: C4 is strongest, solving 6/8 hard tasks with a 33.6\% pass rate and the highest mean AgentQ. Dense feedback alone helps the fixed seed-skill agent but remains limited (C2: 2/8, 12.5\%). Sparse Trace2Skill solves more tasks (C3: 3/8, 27.0\%), while adding dense feedback to the same EA budget solves three additional tasks and raises pass rate by 6.6 percentage points (C4: 6/8, 33.6\%). Thus, C1\(\rightarrow\)C2 isolates dense verifier feedback for the seed-skill agent, C1\(\rightarrow\)C3 isolates task-wise Trace2Skill under sparse final reward, and C3\(\rightarrow\)C4 is the cleanest dense-feedback comparison under matched EA budget. The result supports the central claim that dense feedback is most useful when coupled to task-wise skill evolution rather than treated only as an extra tool in an otherwise fixed seed-skill agent.

\subsection{Quality Metrics}
\begin{figure}[!htbp]
  \centering
  \makebox[\linewidth][c]{\includegraphics[width=1.05\linewidth]{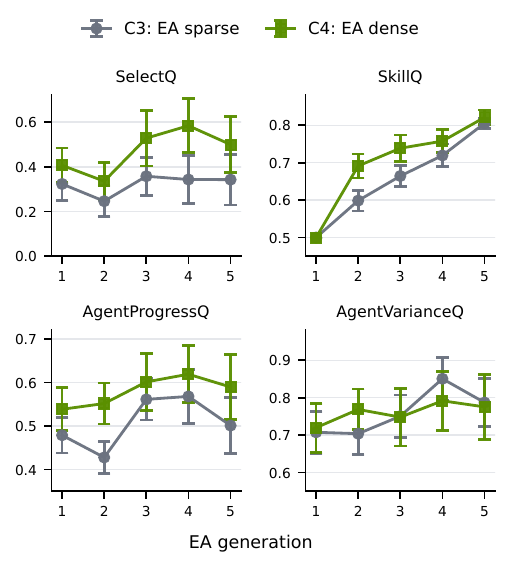}}
  \caption{C3 versus C4 quality dynamics on the 8 hard CVDP tasks. Points show generation-level means; error bars show SEM across valid candidate-skill measurements.}
  \Description{Four-panel line plot comparing sparse and dense Trace2Skill evolution over generations using SelectQ, SkillQ, AgentProgressQ, and AgentVarianceQ.}
  \label{fig:c1-c4-quality}
\end{figure}

Figure~\ref{fig:c1-c4-quality} visualizes artifact-derived dense metrics for C3 and C4, including SelectQ, SkillQ, AgentProgressQ, and AgentVarianceQ across generations. This aggregate view is complementary to the final pass table: SelectQ is the evolutionary survivor score, SkillQ summarizes learned skill content quality, AgentProgressQ summarizes repeat-robust rollout execution quality, and AgentVarianceQ reports repeat stability. The visible uncertainty is expected in long-horizon, multi-turn LLM-agent rollouts: identical skill prompts can still induce different search, edit, and validation trajectories, so repeated samples are used to estimate mean behavior rather than relying on single-run outcomes. We aggregate over valid candidate-skill measurements, excluding invalid or semantic-floor-regressing candidates so the trend reflects viable survivor choices rather than rejected proposals.
Across generations, C4 is higher than C3 in SelectQ and AgentProgressQ, and SkillQ is tied at the shared seed generation before becoming higher for C4 in later generations. This suggests that runtime dense feedback improves both the learned skill content and the agent's execution behavior, rather than only changing the final sparse pass count.

\begin{figure}[!htbp]
  \centering
  \makebox[\linewidth][c]{\includegraphics[width=0.82\linewidth]{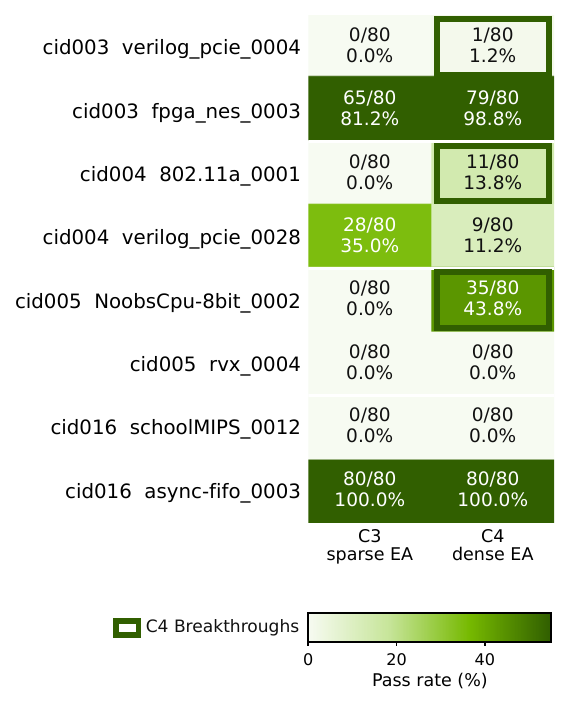}}
  \caption{Task-level C3/C4 verifier outcomes on the 8 hard tasks. Cells show pass count and pass rate over five generations \(\times\) four candidate skills \(\times\) four repeats.}
  \Description{Heatmap with eight task rows and two condition columns, C3 sparse EA and C4 dense EA. Cell color and text show pass rate.}
  \label{fig:c3-c4-cid-outcomes}
\end{figure}

Figure~\ref{fig:c3-c4-cid-outcomes} gives a task-level outcome view that complements the dense metrics. Each row is one hard task, grouped by CID, and each cell aggregates all EA rollouts for that task and condition. The highlighted C4 cells mark breakthrough tasks: C3 obtains no verifier pass, while C4 obtains at least one.

Together, Figures~\ref{fig:c1-c4-quality} and~\ref{fig:c3-c4-cid-outcomes} connect process quality with verifier outcomes. The heatmap shows that C4 is not uniformly better on every task; rather, its value is expanding the set of solvable hard tasks, especially \texttt{verilog\_pcie\_0004}, \texttt{802.11a\_0001}, and \texttt{NoobsCpu-8bit\_0002}. \textbf{The mechanism is skill-agent coadaptation: the verifier signal shapes rollout behavior directly, and the resulting traces give the oracle and mutator clearer lessons for the next generation, forming a closed continuous optimization loop.}

\subsection{Task-Level Pass Rates}
The boxed cells in Figure~\ref{fig:c3-c4-cid-outcomes} are the C4-only breakthroughs. In C3, these tasks had zero verifier passes and no runtime dense-feedback calls. In C4, PCIe reached 1/80 passes, 802.11a reached 11/80, and NoobsCpu reached 35/80. The traces show different levels of dense-feedback dependence. For PCIe, the passing rollout first passed local pulse-merge tests, then feedback exposed hidden functional failures; after further edits, a later feedback call passed the registered \texttt{pulse\_out} variant. For 802.11a, 9 of 11 passing rollouts invoked feedback around \texttt{MAX\_LENGTH=192} Viterbi tests and partial failures; two passing rollouts did not, so the conservative reading is that feedback improved harness alignment and oracle lessons, but was not necessary for every individual pass. For NoobsCpu, the link is strongest: all 35 passing rollouts used feedback, and representative traces show local \texttt{iverilog}/\texttt{vvp} success followed by verifier compile failures, then fixes to include handling, explicit file lists, \texttt{asserts.v}, and \texttt{noobs\_cpu\_defines.vh} until feedback returned pass. Thus, dense feedback converts hidden final failures into actionable in-rollout corrections and clearer oracle/mutator lessons, while still leaving task difficulty and agent stochasticity visible.

\subsection{Limitations}
\begin{itemize}
    \item \textbf{Agent.} Long-horizon agents can take different search, edit, and validation paths under the same skill, so mean metrics need not improve monotonically even when the best survivor skill improves. Future work should better characterize how skill text and dense feedback shape rollout variance and task success.
    \item \textbf{Compute.} Trace2Skill is compute-intensive. A 30-turn horizon gives the agent room to recover, but it also increases rollout variance, verifier traffic, wall-clock time, and token cost. Reducing the required turn budget through better verifier scheduling, earlier stopping, and stronger procedural guidance is therefore a central systems direction.
    \item \textbf{Benchmark.} The OSS CVDP subset studied here is valuable, but it remains concentrated on Verilog-related coding tasks. Broader progress will require the EDA community to contribute open, executable, long-context benchmarks beyond RTL coding so that agentic methods can be compared on synthesis, timing, power, formal verification, and full-flow engineering tasks for the benefit of the wider ecosystem.
\end{itemize}

\section{Conclusion}

Trace2Skill turns execution traces into test-time optimization signal for hardware agents by evolving the natural-language skill injected into a fixed CVDP agent. On eight tasks that failed the seed CVDP agent, including tasks that also failed Claude Code and Codex, the full Trace2Skill configuration with dense verifier feedback solves 6/8 tasks and reaches a 33.6\% hidden-verifier pass rate, compared with 0/8 for the seed agent, 2/8 for dense feedback alone, and 3/8 for sparse skill evolution. The evidence points to skill-agent coadaptation: dense verifier observations sharpen rollout behavior, and the resulting traces give the oracle and mutator better lessons for the next generation. Because the loop requires only tool traces, inspectable skill text, and reliable rewards, it complements RTL-specialized model training and can extend to other verifiable EDA and engineering tasks.

\section*{Acknowledgments}
We thank Joshua Mabry, Arti Jain, and Mark Ho for helpful discussions and feedback.

\bibliographystyle{unsrtnat}
\bibliography{refs}

\clearpage
\onecolumn
\raggedbottom
\appendix

\section{Artifact Appendix}
\newcommand{\AppendixAssetFont}{\fontsize{7.8pt}{8.65pt}\selectfont}

This appendix gives compact, trace-grounded artifacts used by the Trace2Skill pipeline. Full raw traces remain in the experiment directories; hidden verifier collateral, absolute filesystem paths, and user identifiers are omitted.

\subsection*{A1. Task Set and Baseline Outcomes}

Table~\ref{tab:oss24-task-outcomes} gives the exact 24 public OSS-runnable
long-context CVDP tasks used for the matched baseline comparison. Bold rows are
the eight seed-CVDP failures used for the C1--C4 Trace2Skill ablation. The
outcome columns show which of those tasks were also failed by Claude Code and
Codex; failure notes summarize all-agent failures or mark selected seed-only failures.

\begin{table}[H]
\centering
\caption{Public OSS-runnable long-context agentic CVDP tasks used for the
matched baseline comparison. The subset spans the four public CIDs whose
harnesses are runnable without commercial EDA tools. Bold task IDs are the
eight seed-CVDP failures targeted by the C1--C4 ablation study; the outcome
columns show which of them were also failed by Claude Code and Codex, and the final column gives compact failure notes.}
\label{tab:oss24-task-outcomes}
\AppendixAssetFont
\setlength{\tabcolsep}{2.3pt}
\renewcommand{\arraystretch}{1.04}
\begin{tabular}{@{}llp{0.32\textwidth}cccp{0.24\textwidth}@{}}
\toprule
CID & Category & Task ID & \shortstack{CVDP\\Agent} & \shortstack{Claude\\Code} & Codex & Failure note \\
\midrule
cid003 & Spec-to-RTL & \path{cvdp_agentic_heavy_axi4_lite_0002} & \(\surd\) & \(\surd\) & \(\surd\) & -- \\
cid003 & Spec-to-RTL & \path{cvdp_agentic_heavy_corundum_0002} & \(\surd\) & \(\surd\) & \(\surd\) & -- \\
cid003 & Spec-to-RTL & \path{cvdp_agentic_heavy_detect_human_faces_0002} & \(\surd\) & \(\surd\) & \(\times\) & -- \\
cid003 & Spec-to-RTL & \textbf{\texttt{cvdp\_agentic\_heavy\_fpga\_nes\_0003}} & \(\times\) & \(\surd\) & \(\surd\) & CVDP seed failed; frontier agents passed \\
cid003 & Spec-to-RTL & \path{cvdp_agentic_heavy_fpu_wrappers_0002} & \(\surd\) & \(\surd\) & \(\surd\) & -- \\
cid003 & Spec-to-RTL & \textbf{\texttt{cvdp\_agentic\_heavy\_verilog\_pcie\_0004}} & \(\times\) & \(\times\) & \(\times\) & Pulse-merge semantics: \texttt{pulse\_out} high when expected low \\
\addlinespace
cid004 & RTL refinement & \textbf{\texttt{cvdp\_agentic\_heavy\_802.11a\_0001}} & \(\times\) & \(\times\) & \(\times\) & Viterbi decoder semantics: hidden output-bit mismatches \\
cid004 & RTL refinement & \path{cvdp_agentic_heavy_SoomRV_0002} & \(\surd\) & \(\surd\) & \(\times\) & -- \\
cid004 & RTL refinement & \path{cvdp_agentic_heavy_cva6_riscv_core_0005} & \(\times\) & \(\surd\) & \(\times\) & -- \\
cid004 & RTL refinement & \path{cvdp_agentic_heavy_enso_0003} & \(\surd\) & \(\surd\) & \(\surd\) & -- \\
cid004 & RTL refinement & \path{cvdp_agentic_heavy_pulpino_0004} & \(\times\) & \(\times\) & \(\surd\) & -- \\
cid004 & RTL refinement & \textbf{\texttt{cvdp\_agentic\_heavy\_verilog\_pcie\_0028}} & \(\times\) & \(\times\) & \(\times\) & DMA immediate/control-register semantics not preserved \\
\addlinespace
cid005 & Code completion & \textbf{\texttt{cvdp\_agentic\_heavy\_NoobsCpu-8bit\_0002}} & \(\times\) & \(\times\) & \(\times\) & Generated CPU failed Icarus build/elaboration \\
cid005 & Code completion & \path{cvdp_agentic_heavy_cv32e40p_1_0001} & \(\times\) & \(\times\) & \(\times\) & LSU monitor integration failed Icarus build/elaboration \\
cid005 & Code completion & \path{cvdp_agentic_heavy_dual_core_0035} & \(\times\) & \(\times\) & \(\times\) & USB top-level interface/clocking mismatch \\
cid005 & Code completion & \path{cvdp_agentic_heavy_edma_0008} & \(\surd\) & \(\surd\) & \(\surd\) & -- \\
cid005 & Code completion & \textbf{\texttt{cvdp\_agentic\_heavy\_rvx\_0004}} & \(\times\) & \(\times\) & \(\times\) & RVX top-level interface signals missing \\
cid005 & Code completion & \path{cvdp_agentic_heavy_usb_cdc_0002} & \(\times\) & \(\surd\) & \(\surd\) & -- \\
\addlinespace
cid016 & RTL bug fix & \path{cvdp_agentic_heavy_802.11a_0005} & \(\surd\) & \(\times\) & \(\surd\) & -- \\
cid016 & RTL bug fix & \textbf{\texttt{cvdp\_agentic\_heavy\_async-fifo\_0003}} & \(\times\) & \(\surd\) & \(\surd\) & CVDP seed failed; frontier agents passed \\
cid016 & RTL bug fix & \path{cvdp_agentic_heavy_deepfreeze_0027} & \(\times\) & \(\surd\) & \(\surd\) & -- \\
cid016 & RTL bug fix & \path{cvdp_agentic_heavy_enso_0060} & \(\surd\) & \(\times\) & \(\times\) & -- \\
cid016 & RTL bug fix & \textbf{\texttt{cvdp\_agentic\_heavy\_schoolMIPS\_0012}} & \(\times\) & \(\times\) & \(\times\) & Clock-divider and PC-progression semantics failed \\
cid016 & RTL bug fix & \path{cvdp_agentic_heavy_usb_cdc_0017} & \(\times\) & \(\times\) & \(\surd\) & -- \\
\bottomrule
\end{tabular}
\end{table}

\FloatBarrier
\clearpage

\subsection*{A2. Grounded Trace2Skill Generation Examples}

Tables~\ref{tab:app-c4-noobs-generation} and
\ref{tab:app-c4-dense-feedback} compress saved artifacts from completed C4
runs. They are not illustrative mockups: the numbers, tool sequences, and
failure modes are taken from generation metrics, rollout diagnostics, lesson
banks, mutation health logs, and evaluation summaries. The first example shows
a successful dense-feedback generation on \texttt{NoobsCpu-8bit\_0002}; the
second shows how runtime verifier feedback exposed evaluator-relevant failure
information on \texttt{802.11a\_0001}.
\vspace{-0.45em}
\begin{table}[H]
\caption{Trace-grounded C4 generation example from
\texttt{NoobsCpu-8bit\_0002}, generation 5. C4 uses task-wise Trace2Skill plus
runtime dense verifier feedback.}
\label{tab:app-c4-noobs-generation}
\AppendixAssetFont
\setlength{\tabcolsep}{3pt}
\renewcommand{\arraystretch}{1.08}
\begin{tabular}{@{}p{0.18\textwidth}p{0.47\textwidth}p{0.29\textwidth}@{}}
\toprule
Artifact & Recorded trace content & Why it matters \\
\midrule
Run state &
Task \texttt{cvdp\_agentic\_heavy\_NoobsCpu-8bit\_0002}, cid005, C4 dense-feedback EA, generation 5, population 4, repeat 4. The generation produced 9 verifier passes over 16 rollouts; the best individual passed 4/4 repeats. &
The task failed all three baseline agents in Table~\ref{tab:oss24-task-outcomes}, but the evolved C4 lineage reached nonzero and repeatable sparse passes. \\
\addlinespace
Selected survivor &
The selected child was \texttt{child\_gen5\_1}. Its logged values were SelectQ 1.104, sparse pass@1 1.000, robust utility 0.853, SkillQ 0.894, and AgentProgressQ 0.820. The selected skill had no missing critical directive, no contradiction, and no invalid-skill propagation. &
Selection is not based on a single lucky rollout alone: it combines sparse success, robust repeated progress, and skill quality while enforcing validity checks. \\
\addlinespace
Passing rollout trace &
One passing rollout, \texttt{ind1\_r0}, made 30 tool calls. The compact sequence was: read skill and files, inspect modules, write \path{vmodel/noobs_cpu.v}, run \texttt{iverilog} and \texttt{vvp}, call \texttt{verify\_feedback}, edit again, repeat compile/simulate/feedback twice, and finish with a hidden-verifier pass. &
The trace shows the intended behavior of dense feedback: the agent uses local compile/simulation and bounded black-box verifier feedback during the rollout, then final hidden verification remains the official reward. \\
\addlinespace
Progress components &
Generation-level component means were edit quality 0.956, efficiency 0.869, execution phase 0.956, harness alignment 0.872, path grounding 0.403, and verifier partial 0.563. Path-grounding diagnostics still recorded shadow-path issues between prompt-discovered files and same-basename files under \path{vmodel/}. &
The metric breakdown is useful because a sparse pass alone would hide both the successful behavior and the remaining weakness. Here, C4 solved the task while still exposing a path-grounding risk to analyze. \\
\addlinespace
Oracle lessons &
The cumulative lesson bank preserved concrete rules: read submodules before wiring, use tight \texttt{iverilog} $\rightarrow$ \texttt{vvp} loops, compile \path{asserts.v} exactly once, prefer stable \texttt{-I} include roots over include churn, and use \texttt{verify\_feedback} or build artifacts to infer expected wiring and instance names. &
The oracle converts repeated rollout traces into task-local procedural memory instead of generic hardware advice. \\
\addlinespace
Evolved skill excerpt &
The generation-5 skill explicitly allowed \texttt{verify\_feedback}, instructed the agent to inspect \texttt{ifetch}, \texttt{idecode}, \texttt{register\_file}, and \texttt{execute}, warned about duplicate module/include handling, and required compile/run cycles after small edits. &
The injected skill is an auditable artifact. It preserves durable lessons while adding task-specific guidance learned from earlier generations. \\
\bottomrule
\end{tabular}
\end{table}

\begin{table}[H]
\caption{Runtime dense-verifier feedback and mutation-repair evidence from
\texttt{802.11a\_0001}.}
\label{tab:app-c4-dense-feedback}
\AppendixAssetFont
\setlength{\tabcolsep}{3pt}
\renewcommand{\arraystretch}{1.08}
\begin{tabular}{@{}p{0.18\textwidth}p{0.47\textwidth}p{0.29\textwidth}@{}}
\toprule
Artifact & Recorded trace content & Why it matters \\
\midrule
Dense feedback call &
A generation-4 rollout edited \path{Hardware/Receiver/ViterbiDecoder/ViterbiDecoder.v}, compiled it, ran local simulation, then called \texttt{verify\_feedback}. The saved diagnostics record final mode \texttt{fail:simulation\_or\_assertion}, phase \texttt{P4-tested}, Docker exit code 1, and a sanitized verifier tail at \texttt{MAX\_LENGTH=192} with \texttt{SystemExit: ERROR: Failed 3 of 5 tests}. Other generation-4 diagnostics record 3/5 and 5/5 failures, while successful rollouts report \texttt{TESTS=5 PASS=5 FAIL=0}. &
The agent receives evaluator-relevant functional observations without seeing the hidden harness or reference solution. This is denser than a final binary failure but still black-box. \\
\addlinespace
Agent response &
The saved generation prompts and responses explicitly discuss the \texttt{MAX\_LENGTH=192} setting, 3/5 or 5/5 official-runner failures, 192-bit output mismatches, and timing or traceback behavior. After the feedback call, the sampled rollout continued with RTL edits and additional \texttt{iverilog}/\texttt{vvp} checks. &
The trace confirms that the dense tool affected the rollout trajectory rather than merely appearing in the tool list. \\
\addlinespace
Oracle diagnosis &
The generation summary identified a harmful pattern: many failed rollouts relied on self-authored benches such as \path{ViterbiDecoder_tb2.v}, \path{ViterbiDecoder_tb3.v}, \path{mini_tb.v}, or \path{debug_tb.v}, causing timeouts or mismatch with harness parameters. The successful rollouts aligned better with the official runner and \texttt{verify\_feedback} evidence. &
Dense feedback helps the oracle distinguish locally plausible behavior from hidden-verifier-aligned behavior, producing sharper lessons for the next mutation. \\
\addlinespace
Mutation repair &
Mutation-health logs show repaired children rather than blind rejection. A generation-3 repaired child had sanitizer repairs \texttt{rewrote\_harness\_execution\_to\_visible\_metadata} and \texttt{rewrote\_unconfirmed\_path\_guidance}; its integration coverage was 1.000 versus parent coverage 0.625, with zero missing critical directives and zero contradictions after repair. &
Useful semantic proposals are salvaged when possible, while hidden-harness leakage, ungrounded path guidance, or contradictory instructions are removed before the skill enters the tournament. \\
\bottomrule
\end{tabular}
\end{table}
\FloatBarrier
\clearpage

\subsection*{A3. Abbreviated Raw EA Generation Trace}

Table~\ref{tab:app-raw-ea-generation} records one literal, abbreviated
generation transition from the saved C4 artifacts. It follows the same
task-local chain used by the implementation: task prompt, rollout trace,
scoring, oracle summary, mutator health, and the next injected skill. Long tool
lists are shortened, and hidden verifier paths or source are omitted.

\begin{table}[H]
\caption{Abbreviated raw artifacts for one C4 EA transition on
\texttt{NoobsCpu-8bit\_0002}, generation 4 to generation 5.}
\label{tab:app-raw-ea-generation}
\AppendixAssetFont
\setlength{\tabcolsep}{3pt}
\renewcommand{\arraystretch}{1.05}
\begin{tabular}{@{}>{\raggedright\arraybackslash}p{0.15\textwidth}>{\raggedright\arraybackslash}p{0.79\textwidth}@{}}
\toprule
Stage & Abbreviated raw artifact text \\
\midrule
Task prompt &
\texttt{Create a top-level RTL module named noobs\_cpu in /code/vmodel/noobs\_cpu.v ...}
Subblocks listed by the task: \texttt{ifetch}, \texttt{idecode},
\texttt{reg\_file}, and \texttt{execute}; headers:
\texttt{asserts.v} and \texttt{noobs\_cpu\_defines.vh}. Deliverable:
\texttt{iverilog -g2012 noobs\_cpu.v ifetch.v idecode.v register\_file.v execute.v}. \\
\addlinespace
Rollout trace &
\texttt{rollout\_diagnostics.jsonl: gen4 ind2 r0 outcome=pass reward=1.0 phase=P4-tested num\_tool\_calls=28}.
Tool sequence:
\texttt{[cat, ls, ..., echo, rg, iverilog, vvp, verify\_feedback, edit, edit, iverilog, vvp, verify\_feedback, edit, iverilog, vvp, verify\_feedback, show\_changes]}.
Files written included \path{vmodel/noobs_cpu.v}. Hidden-verifier tail:
\texttt{test\_cpu.cpu\_smoke\_test PASS; TESTS=1 PASS=1 FAIL=0; 1 passed in 0.14s}. \\
\addlinespace
Scoring &
\texttt{combined\_selection\_fitness.json: ind2 child\_gen4\_2 pass@1=0.75 SelectQ=0.818482 robust\_utility=0.559034 SkillQ=0.596722 AgentBehaviorQ=0.481760 progress\_lcb95=0.432600 progress\_mean=0.678400}.
Other candidates had SelectQ \texttt{0.556877}, \texttt{0.294436}, and
\texttt{0.050071}; \texttt{child\_gen4\_2} survived. \\
\addlinespace
Oracle summary &
\texttt{oracle\_feedback.md: best=0.818 mean=0.430 worst=0.050 rollouts=16 (6 pass, 10 fail)}.
\texttt{[cid005] NoobsCpu-8bit\_0002 avg=38\% (75\%, 50\%, 25\%, 0\%) [SOMETIMES SOLVED]}.
KEEP: read submodules before wiring; use tight \texttt{iverilog -> vvp}
cycles; compile \path{asserts.v} exactly once. ADD: protected path grounding;
coordinate top/submodule fixes. REMOVE: prefer editing same-basename
\path{vmodel/} mirrors; remove or guard asserts. \\
\addlinespace
Mutator and repair &
\texttt{mutation\_handoff.md: gen4\_ind2 EFFECTIVE selection +0.523 pass +0.500 progress +0.242}.
\texttt{mutation\_health.jsonl child\_slot=1 parent=child\_gen4\_2 ok=true has\_oracle\_feedback=true similarity=0.468 coverage=1.000 parent\_coverage=0.792 missing\_critical=0 parent\_missing=3 remove\_violation=0 contradiction=0 repair\_attempted=false}. \\
\addlinespace
Next skill &
\texttt{ea\_survivor\_skill.md: Hard Rules: use only ... verify\_feedback; text-only answers ignored; compile then simulate. Protected Path Grounding: edit and validate exact target files ... Do NOT write or validate only same-basename shadow paths. Iverilog Semantics: -I only affects include lookup; pass all required .v files exactly once. Asserts Handling: compile asserts.v exactly once; do not remove asserts.} \\
\bottomrule
\end{tabular}
\end{table}
\FloatBarrier
\clearpage

\subsection*{A4. Execution Contracts, Safety, and Telemetry}

Algorithm~\ref{alg:appendix-parent-child-flow} gives the compact generation update
implemented by the current EA loop. The parent survivor is always included in
the next population, children are repaired before evaluation, and the oracle
summarizes the full task-local population so failed children can contribute
negative lessons.

\begin{algorithm}[H]
\caption{Task-wise Trace2Skill evolutionary update implemented by the standard EA loop.}
\label{alg:appendix-parent-child-flow}
\AppendixAssetFont
\begin{algorithmic}[1]
\Require Task $x$; population size $K$; repeats $R$; generation-$g$ population $P_g=[S_0,\ldots,S_{K-1}]$, where $S_0$ is the carried parent or seed skill; previous lesson bank $L_{g-1}$.
\Ensure Unevaluated population $P_{g+1}$ for the next generation and updated lesson bank $L_g$.
\For{$i=0$ to $K-1$}
    \State $T_i \gets \{\Call{RolloutAndVerify}{x,S_i,r}\}_{r=1}^{R}$
    \State $M_i \gets \Call{ComputeMetrics}{T_i,S_i}$
    \State $q_i \gets \Call{SelectQ}{M_i,S_i,S_0}$
\EndFor
\State $S_g^{\star} \gets \arg\max_{S_i \in P_g} q_i$ \Comment{survivor selected only after rollouts}
\State \Call{WriteArtifacts}{$\{T_i,M_i,q_i\}_{i=0}^{K-1},S_g^{\star}$}
\State $O_g \gets \Call{Oracle}{\{T_i\}_{i=0}^{K-1},S_g^{\star},L_{g-1}}$ \Comment{summarize all population evidence}
\State $L_g \gets \Call{UpdateLessonBank}{L_{g-1},O_g,\{M_i\}_{i=0}^{K-1}}$
\State $H_g \gets \Call{BuildMutationHandoff}{O_g,L_g}$
\State $P_{g+1} \gets [\Call{CarryForReevaluation}{S_g^{\star}}]$
\While{$|P_{g+1}| < K$}
    \State $\widehat{S} \gets \Call{Mutator}{S_g^{\star},H_g}$
    \State $\widetilde{S} \gets \Call{RepairAndSanitize}{\widehat{S}}$
    \If{\Call{MutationHealthOK}{$\widetilde{S},S_g^{\star}$}}
        \State append $\widetilde{S}$ to $P_{g+1}$ \Comment{evaluated in generation $g+1$}
    \Else
        \State append \Call{CarryForReevaluation}{$S_g^{\star}$} to $P_{g+1}$ \Comment{parent-preserving fallback}
    \EndIf
\EndWhile
\State \Return $P_{g+1},L_g$
\Function{SelectQ}{$M_i,S_i,S_0$}
    \State $u_i \gets 0.60F_{\mathrm{LCB}} + 0.20\bar F_{\mathrm{progress}} + 0.20Q_{\mathrm{skill}}$
    \If{\Call{Invalid}{$S_i$} \textbf{or} $Q_{\mathrm{skill}}(S_i)+\tau < Q_{\mathrm{skill}}(S_0)$}
        \State \Return $-1$
    \EndIf
    \State \Return $\mathrm{pass@1}_i + \epsilon_i\,\mathrm{clip}(u_i,0,1)$
\EndFunction
\end{algorithmic}
\vspace{2pt}
\noindent\textit{Implementation note.} The code carries the selected survivor into the next generation for re-evaluation, mutates that same survivor to form children, repairs and sanitizes proposed children before injection, and defers child selection until those children have their own rollouts. Thus no unevaluated child can become the next parent purely from a semantic comparison.
\end{algorithm}
\FloatBarrier
\clearpage

\begin{table}[H]
\caption{Visibility and dense-feedback contracts checked by saved
artifacts.}
\label{tab:app-safety-contracts}
\AppendixAssetFont
\setlength{\tabcolsep}{3pt}
\renewcommand{\arraystretch}{1.08}
\begin{tabular}{@{}p{0.18\textwidth}p{0.45\textwidth}p{0.31\textwidth}@{}}
\toprule
Contract & Grounded artifact evidence & Agent-visible effect \\
\midrule
Hidden verifier isolation &
Execution contracts mark the hidden verifier as unavailable to the live rollout. Dense feedback runs on a private copy; hidden harness files, raw cocotb source, Docker compose files, reference patches, and hidden paths are not written into the live agent sandbox. &
The agent can use \texttt{iverilog}, \texttt{vvp}, and, in dense configurations, \texttt{verify\_feedback}; it cannot inspect or edit the hidden evaluator. \\
\addlinespace
Sanitized dense feedback &
Dense-feedback contracts and telemetry record bounded summaries such as pass/fail or final mode, compile/simulation status, timeout/failure phase, partial test information when available, and sanitized failure context. Raw hidden harness source and reference solutions are omitted. &
Feedback resembles black-box functional test feedback: enough to guide the next RTL edit, not enough to reveal the answer. \\
\addlinespace
Mutation sanitizer &
Mutation-health logs record allowed tools, visible-context path counts, hidden-verifier status, coverage against oracle directives, missing critical directives, contradictions, and repair actions such as rewriting hidden-harness execution guidance into visible metadata guidance. &
Good but defective proposals can be repaired and evaluated; unsafe or ungrounded content is prevented from propagating silently. \\
\bottomrule
\end{tabular}
\end{table}
\clearpage

\begin{table}[H]
\caption{Telemetry artifacts retained for post hoc analysis.}
\label{tab:app-artifact-map}
\AppendixAssetFont
\setlength{\tabcolsep}{3pt}
\renewcommand{\arraystretch}{1.08}
\begin{tabular}{@{}p{0.23\textwidth}p{0.39\textwidth}p{0.30\textwidth}@{}}
\toprule
Artifact family & Representative files & Analysis enabled \\
\midrule
Run state and contracts &
\path{status.json}, \path{run_config.json}, \path{execution_contract.json}, \path{preflight_report.json} &
Reconstruct branch/configuration, task, generation, population, repeat budget, model choices, tool visibility, and hidden-verifier policy. \\
\addlinespace
Rollout traces &
\path{rollout_diagnostics.jsonl}, per-individual \path{eval_summary_*.json}, \path{rollouts_*_agent_metrics.json} &
Recover tool sequences, files read/written, compile commands, verifier calls, final pass/fail, path-grounding issues, and phase reached. \\
\addlinespace
Dense metrics and selection &
\path{generation_metrics.json}, \path{combined_selection_fitness.json}, \path{rollout_model_check_*.json} &
Compute SelectQ, SkillQ, AgentProgressQ, AgentVarianceQ, pass@1, pass@k, confidence intervals, and selected-survivor rationale. \\
\addlinespace
Learning artifacts &
\path{oracle_feedback.md}, \path{lesson_bank.md}, \path{mutation_handoff.md}, \path{skill_patch_plan.json}, \path{best_skill.md}, \path{ea_survivor_skill.md} &
Audit whether rollout failures were summarized into concrete lessons and whether those lessons survived into the next injected skill. \\
\addlinespace
Safety and repair &
\path{skill_sanitization.jsonl}, \path{skill_integration.jsonl}, \path{mutation_health.jsonl} &
Check hidden-harness leakage prevention, unavailable-tool removal, contradiction repair, and critical-guidance preservation. \\
\bottomrule
\end{tabular}
\end{table}

\clearpage
\subsection*{A5. Repository Component Map}

The implementation is spread across a research orchestrator, a vendored NeMo
Gym CVDP harness, skill prompts, optimizer modules, and run artifacts. The
closed-loop path is: configuration and task JSONL $\rightarrow$ seed or evolved
skill $\rightarrow$ NeMo Gym rollout servers $\rightarrow$ tool and verifier
traces $\rightarrow$ dense metrics and SelectQ $\rightarrow$ oracle summary
$\rightarrow$ mutator and repair $\rightarrow$ next-generation skill.
Figure~\ref{fig:app-repo-tree} gives a tree-like repo view, while
Table~\ref{tab:app-repo-component-map} maps the same components to their
pipeline roles. Generated caches and legacy configs are omitted from the tree.

\begin{figure}[H]
\caption{Tree view of the repo-relative Trace2Skill implementation and
representative run artifacts.}
\label{fig:app-repo-tree}
\Description{Tree view listing the repository directories, core pipeline files, tests, and representative run-artifact files used by Trace2Skill.}
\vspace{-0.45em}
\AppendixAssetFont
\begin{verbatim}
trace-skill/
|-- orchestration/
|   |-- run_experiment.py
|   `-- configs/mlcad_per_task_oss24_ea/
|       |-- manifest.json
|       `-- cvdp_per_task_*.yml
|-- nemogym/
|   |-- data/cid_splits_manifest.json
|   |-- nemo_gym/cli.py
|   |-- responses_api_agents/cvdp_agentic_heavy_agent/app.py
|   |-- responses_api_models/openai_model/app.py
|   `-- resources_servers/cvdp_agentic_heavy/
|       |-- app.py
|       `-- context_pruning.py
|-- skills/
|   |-- rollout/*.md
|   |-- optimizer/*.md
|   `-- analysis/*.md
|-- optimizer/
|   |-- run_optimizer.py
|   |-- fitness.py
|   |-- skill_selector.py
|   |-- streaming_oracle.py
|   |-- prompt_mutator.py
|   |-- skill_sanitizer.py
|   |-- skill_integration.py
|   |-- lesson_bank.py
|   `-- telemetry.py
|-- eval/
|   |-- convert_rollouts.py
|   |-- run_analyzer.py
|   `-- run_judge.py
|-- tests/
|   |-- test_standard_ea_flow.py
|   |-- test_dense_progress_fitness.py
|   |-- test_skill_sanitizer.py
|   |-- test_skill_integration.py
|   |-- test_mutation_health.py
|   `-- test_traceability_artifacts.py
`-- paper/mlcad2026_trace2skill/
    `-- appendix/artifact_appendix.tex

<artifact_root>/trace_skill_runs/.../
`-- runs/<task_run>/
    |-- status.json
    |-- run_config.json
    |-- execution_contract.json
    `-- gen*/
        |-- rollout_diagnostics.jsonl
        |-- generation_metrics.json
        |-- combined_selection_fitness.json
        |-- oracle_feedback.md
        |-- mutation_handoff.md (mutation generations)
        |-- lesson_bank.md
        `-- ea_survivor_skill.md
\end{verbatim}
\vspace{-0.9em}
\end{figure}

\begin{table}[H]
\caption{Repo-relative components used by the end-to-end Trace2Skill pipeline.}
\label{tab:app-repo-component-map}
\AppendixAssetFont
\setlength{\tabcolsep}{3pt}
\renewcommand{\arraystretch}{1.08}
\begin{tabular}{@{}>{\raggedright\arraybackslash}p{0.18\textwidth}>{\raggedright\arraybackslash}p{0.35\textwidth}>{\raggedright\arraybackslash}p{0.39\textwidth}@{}}
\toprule
Layer & Repo or artifact components & Role in the pipeline \\
\midrule
Experiment orchestration &
\path{orchestration/configs/mlcad_per_task_oss24_ea/manifest.json}; per-task YAML files; \path{orchestration/run_experiment.py}. &
Defines C1--C4 condition, task list, population, generations, repeats, models, output directory, and dense-feedback flag. \\
\addlinespace
CVDP task data &
\path{nemogym/data/cid_splits_manifest.json}; CID folders under \path{nemogym/data/}; run-local \path{tasks/*.jsonl}. &
Stores the visible task prompt, repository context, allowed tools, seed-skill reference, and OSS-pruned task materialization. Hidden verifier collateral is not included in the live agent workspace. \\
\addlinespace
NeMo Gym runtime &
\path{nemogym/nemo_gym/cli.py}; \path{nemogym/responses_api_agents/cvdp_agentic_heavy_agent/app.py}; \path{nemogym/responses_api_models/openai_model/app.py}. &
Starts the rollout agent and model-serving endpoints. The CVDP agent loop receives the skill, calls visible tools, edits RTL, and records the action trajectory. \\
\addlinespace
CVDP resource server &
\path{nemogym/resources_servers/cvdp_agentic_heavy/app.py}; \path{context_pruning.py}; CVDP conversion scripts. &
Implements file tools, \texttt{iverilog}, \texttt{vvp}, final \texttt{verify}, and optional \texttt{verify\_feedback}. Dense feedback runs the hidden verifier on a private copy and returns only sanitized functional observations. \\
\addlinespace
Skill and prompt assets &
\path{skills/rollout/*.md}; \path{skills/optimizer/*.md}; \path{skills/analysis/*.md}. &
Rollout skills provide CID/task guidance. Optimizer skills define oracle, mutation, repair, and handoff prompts. Analysis skills support offline trajectory interpretation. \\
\addlinespace
Trace2Skill optimizer &
\path{optimizer/run_optimizer.py}; \path{fitness.py}; \path{skill_selector.py}; \path{streaming_oracle.py}; \path{prompt_mutator.py}; \path{skill_sanitizer.py}; \path{skill_integration.py}; \path{lesson_bank.py}; \path{telemetry.py}. &
Evaluates candidate skills, computes dense metrics and SelectQ, selects the survivor, summarizes population traces, mutates children, repairs unsafe proposals, updates the lesson bank, and writes traceability artifacts. \\
\addlinespace
NAT-style conversion and analysis &
\path{eval/convert_rollouts.py}; \path{eval/run_analyzer.py}; \path{eval/run_judge.py}. &
Converts raw rollout JSONL into structured trajectories and behavior metrics using NAT-style evaluation patterns. The current repo does not vendor or invoke the NAT package directly. \\
\addlinespace
Run artifacts &
Run directories with \path{status.json}, \path{run_config.json}, \path{execution_contract.json}, \path{gen*/rollout_diagnostics.jsonl}, \path{generation_metrics.json}, \path{combined_selection_fitness.json}, \path{oracle_feedback.md}, \path{mutation_handoff.md}, \path{lesson_bank.md}, and \path{ea_survivor_skill.md}. &
Provide the evidence used for post hoc analysis, paper figures, pass-rate tables, dense-metric plots, skill lineage inspection, safety audits, and C4 breakthrough trace explanations. \\
\addlinespace
Validation tests &
\path{tests/test_standard_ea_flow.py}; \path{test_dense_progress_fitness.py}; \path{test_skill_sanitizer.py}; \path{test_skill_integration.py}; \path{test_mutation_health.py}; \path{test_traceability_artifacts.py}; verifier-feedback smoke artifacts. &
Check that the EA loop, scoring, sanitizer, proposal repair, telemetry, prompt hygiene, and dense-feedback policy behave as intended before launching long runs. \\
\bottomrule
\end{tabular}
\end{table}

\clearpage
\subsection*{A6. NeMo Gym Rollout Sequence}

Figure~\ref{fig:app-nemogym-rollout} expands the rollout-server interaction
used by the CVDP agent. The head server launches a run, the agent alternates
between model generations and resource-server tool execution, and the resource
server returns both intermediate tool results and final verifier rewards.

\begin{figure}[H]
\caption{NeMo Gym rollout sequence for one submitted run.}
\label{fig:app-nemogym-rollout}
\Description{Sequence diagram showing the client fetching configuration, launching a run, the agent interacting with the model and resource server, optional tool calls, verification, scoring, and returned rewards.}
\centering
\vspace{-0.25em}
\resizebox{0.98\textwidth}{!}{%
\begin{tikzpicture}[
    x=1cm,y=1cm,
    >=Stealth,
    font=\scriptsize,
    participant/.style={draw=violet!18, fill=violet!8, rounded corners=1.5pt, minimum width=2.1cm, minimum height=0.58cm, font=\bfseries\scriptsize},
    lifeline/.style={draw=violet!20, line width=0.6pt},
    msg/.style={->, line width=0.55pt, draw=black!68},
    ret/.style={->, dashed, line width=0.55pt, draw=black!58},
    region/.style={draw=violet!22, dashed, rounded corners=2pt, line width=0.6pt},
    tag/.style={draw=violet!16, fill=violet!7, rounded corners=1pt, inner xsep=3pt, inner ysep=1.2pt, font=\scriptsize},
    num/.style={circle, fill=black!78, text=white, inner sep=1.2pt, font=\tiny\bfseries}
]
\foreach \name/\x in {Client/0,Head Server/3.2,Agent/6.4,Model/9.6,Resources/12.8} {
    \node[participant] at (\x,0) {\name};
    \node[participant] at (\x,-11.65) {\name};
    \draw[lifeline] (\x,-0.32) -- (\x,-11.33);
}
\draw[region] (5.85,-2.45) rectangle (13.20,-10.75);
\node[tag, anchor=west] at (5.95,-2.45) {loop};
\node[anchor=north] at (9.55,-2.52) {[Each prompt in parallel]};
\draw[region] (6.05,-4.05) rectangle (13.00,-9.80);
\node[tag, anchor=west] at (6.15,-4.05) {loop};
\node[anchor=north] at (9.65,-4.12) {[Generation loop]};
\draw[region] (6.25,-5.80) rectangle (12.95,-7.75);
\node[tag, anchor=west] at (6.35,-5.80) {opt};
\node[anchor=north] at (9.75,-5.87) {[Tool calls]};

\draw[msg] (0,-1.05) -- node[above]{\texttt{GET /global\_config\_dict\_yaml}} (3.2,-1.05);
\node[num] at (0.02,-1.05) {1};
\draw[ret] (3.2,-1.55) -- node[above]{Server addresses} (0,-1.55);
\node[num] at (3.2,-1.55) {2};
\draw[msg] (0,-2.15) -- node[above]{\texttt{POST /run}} (6.4,-2.15);
\node[num] at (0.02,-2.15) {3};

\draw[msg] (6.4,-3.25) -- node[above]{Seed session} (12.8,-3.25);
\node[num] at (6.4,-3.25) {4};
\draw[ret] (12.8,-3.75) -- node[above]{OK} (6.4,-3.75);
\node[num] at (12.8,-3.75) {5};

\draw[msg] (6.4,-5.02) -- node[above]{Generate response} (9.6,-5.02);
\node[num] at (6.4,-5.02) {6};
\draw[ret] (9.6,-5.52) -- node[above]{Output} (6.4,-5.52);
\node[num] at (9.6,-5.52) {7};

\draw[msg] (6.4,-6.82) -- node[above]{Execute tool} (12.8,-6.82);
\node[num] at (6.4,-6.82) {8};
\draw[ret] (12.8,-7.32) -- node[above]{Result} (6.4,-7.32);
\node[num] at (12.8,-7.32) {9};

\draw[msg] (6.4,-8.65) -- node[above]{Verify and score} (12.8,-8.65);
\node[num] at (6.4,-8.65) {10};
\draw[ret] (12.8,-9.15) -- node[above]{Reward} (6.4,-9.15);
\node[num] at (12.8,-9.15) {11};

\draw[ret] (6.4,-10.65) -- node[above]{Results with rewards} (0,-10.65);
\node[num] at (6.4,-10.65) {12};
\end{tikzpicture}%
}
\vspace{-0.4em}
\end{figure}

\end{document}